%% file: acl22-clickbait-spoiling-frame.tex
\pdfoutput=1

\PassOptionsToPackage{hyphens}{url}
\documentclass[11pt]{article}

\usepackage{acl}

\usepackage{times}
\usepackage{latexsym}

\usepackage[T1]{fontenc}

\usepackage[utf8]{inputenc}

\usepackage{microtype}

\usepackage[german,american]{babel}
\newcommand{\Ni}{(1)~}
\newcommand{\Nii}{(2)~}
\newcommand{\Niii}{(3)~}

\usepackage{booktabs} 
\usepackage{multirow}
\usepackage{graphicx}
\DeclareGraphicsExtensions{.pdf,.ai,.jpg,.png}
\setkeys{Gin}{pagebox=artbox}
\graphicspath{{./}{./acl22-clickbait-spoiling-figures/}}

\newcommand{\bsfigure}[3][scale=1.0]{%
  \begin{figure}[tb]
    \centering
    \includegraphics[#1]{#2}
    \vspace{-3ex}
    \caption{#3}\label{#2}
    \vspace{-2ex}
  \end{figure}}

\RequirePackage{type1cm}
\RequirePackage{soul}
\setstcolor{blue}

\newcommand{\bscom}[3][]{%
  \noindent
  \st{#2}{\color{blue}\fontsize{8}{8}\selectfont\,#3}%
  \ifx\\#1\\\else{\color{violet}\fontsize{8}{8}\selectfont\,[#1]}\fi}

\newcommand{\tf}{\ensuremath{\mathit{tf}}}
\newcommand{\idf}{\ensuremath{\mathit{idf}}}
\newcommand{\tfidf}{\ensuremath{\mathit{tf}\kern-0.15em\cdot\kern-0.15em\mathit{idf}}}

\defcitealias{koechley:2012}{Peter Koechley}

\begin{document}

\input{acl22-clickbait-spoiling-pre}
\input{acl22-clickbait-spoiling-part1}
\input{acl22-clickbait-spoiling-part2}
\input{acl22-clickbait-spoiling-part3}
\input{acl22-clickbait-spoiling-part4}
\input{acl22-clickbait-spoiling-part5}
\input{acl22-clickbait-spoiling-part6}
\input{acl22-clickbait-spoiling-sum}
\input{acl22-clickbait-spoiling-ethics}

\bibliographystyle{acl_natbib}
\bibliography{acl22-clickbait-spoiling-lit}

\end{document}

%% file: acl22-clickbait-spoiling-pre.tex
\title{Clickbait Spoiling via Question Answering and Passage Retrieval}

\author{Matthias Hagen$^{1}$ \\
\And
Maik Fr{\"o}be$^{1}$
\And
Artur Jurk$^{1}$ \\
\And
Martin Potthast$^{2}$ \\
\AND
\\[-5ex]
$^{1}$Martin-Luther-Universit{\"a}t Halle-Wittenberg \\
$^{2}$Leipzig University
}

\date{}

\maketitle

\begin{abstract}
We introduce and study the task of clickbait spoiling: generating a short text that satisfies the curiosity induced by a clickbait post. Clickbait links to a web page and advertises its contents by arousing curiosity instead of providing an informative summary. Our contributions are approaches to classify the type of spoiler needed (i.e., a phrase or a passage), and to generate appropriate spoilers. A large-scale evaluation and error analysis on a new corpus of 5,000~manually spoiled clickbait posts---the Webis Clickbait Spoiling Corpus~2022---shows that our spoiler type classifier achieves an accuracy of~80\%, while the question answering model DeBERTa-large outperforms all others in generating spoilers for both types.
\end{abstract}

%% file: acl22-clickbait-spoiling-part1.tex
\section{Introduction}

Clickbait is the term used to describe posts in social media that are intended to inappropriately entice their readers to visit a web page. This is achieved through formulations such as sensationalism or cataphors that are believed to create a so-called curiosity gap: ``a form of cognitively induced deprivation that arises from the perception of a gap in knowledge or understanding'' \cite{loewenstein:1994}. Clickbait is perceived as inappropriate since its resolution is usually ordinary or trivial, comprising little more than a phrase, short passage, or a list of things that could just as easily have been included in the post. This observation motivates us to introduce the task of clickbait spoiling: identifying or generating a spoiler for a clickbait post.

Figure~\ref{twitter-clickbait-examples} shows four examples of clickbait on Twitter, along with spoilers. The first two tweets explicitly or implicitly promise a surprising resolution to spark curiosity, but their spoilers are brief and trivial. The linked page of the first tweet adds almost nothing, and the spoiler of the second is common sense. The third spoiler is a passage from the linked page, and the fourth is a list of things. Even though there are length limits to the informativeness of tweets, the spoilers in all examples could easily have been part of the original tweets.

\bsfigure[width=\linewidth]{twitter-clickbait-examples}{Examples of clickbait tweets and spoilers for them extracted from the respective linked web page.}

This paper reports about our investigation into clickbait spoiling and the following contributions:
\Ni
The Webis Clickbait Spoiling Corpus~2022 (Webis-Clickbait-22), consisting of 5,000~clickbait posts, their linked pages and a spoiling piece of text therein.%
\footnote{Data: \url{https://webis.de/data.html?q=clickbait}}
\Nii
A two-step approach to clickbait spoiling that first classifies a clickbait post according to its spoiler type (phrase or passage), and then treats spoiling either as a question answering or as a passage retrieval task.
\Niii
A systematic evaluation of state-of-the-art methods for spoiler type classification, question answering, and passage retrieval.%
\footnote{Code: \url{https://github.com/webis-de/ACL-22}}
Although the first step of spoiler type classification is not necessary, our results suggest that it can be helpful. Even more so, as we have not yet tackled multipart spoilers (bottom example in Figure~\ref{twitter-clickbait-examples}; 876~cases also part of our corpus) that probably require a different spoiling approach.

%% file: acl22-clickbait-spoiling-part2.tex
\section{Related Work}
\label{related-work}

Following an overview of research on clickbait and its operationalization so far, models of question answering and passage retrieval are examined.

\subsection{Clickbait and its Operationalization}

The underlying assumption of most research on clickbait is that it is a form of data-driven optimization of social media posts to exploit the curiosity gap described by \citet{loewenstein:1994}. At least that's what \citetalias{koechley:2012}~\citeyearpar{koechley:2012}, the CEO of Upworthy, claimed. Upworthy became one of the first major spreaders of clickbait on Facebook, and their success has prompted Facebook to change its news recommendation algorithms to curb the amount of clickbait, twice \cite{elarini:2014,peysakhovich:2016}.

Exploratory and theoretical studies of clickbait and its impact on journalism analyzed its prevalence for more than 150~publishers \cite{rony:2017}; its economics for the news market \cite{munger:2020}; its impact on perceptions of credibility and quality (overall negative) \cite{molyneux:2020}; and noted a slow decline over the past decade \cite{lischka:2021}.

Journalistic studies of this kind rely on clickbait detection technologies. Originally proposed by \citet{rubin:2015a} but not followed up, \citet{potthast:2016b} and \citet{chakraborty:2016} independently developed the first detectors. Starting from a shared task organized by \citet{potthast:2018w} shortly after, more than 50~approaches have been contributed to date. An overview is beyond the scope of our work, but transformer models dominate this task as well. For the clickbait generation task, preceded by a rule-based generator \cite{eidnes:2015}, only \citet{shu:2018} and \citet{xu:2019} have presented more advanced models, while \citet{karn:2019} generate teaser headlines that are explicitly not meant to be clickbait. So far, no attempt has been made to generate spoilers for clickbait.

\subsection{Question Answering}
\label{question-answering}

If one considers clickbait spoiling as a question answering problem, there are numerous possible solutions. Among the available question-answering benchmarks \cite{dzendzik:2021}, we select two to choose appropriate state-of-the-art models for our evaluation:
\Ni
SQuAD \cite{rajpurkar:2016} compiles 107,785 questions and answers based on 536~Wikipedia articles. Although a wide range of questions and answers are included, the vast majority of~93.6\% are factual (32\%~names, 31.8\%~noun phrases, 19.8\%~numbers, 5.5\%~verb phrases, and 3.9\%~adjective phrases), while the remainder are descriptive (3.7\%~clauses and 2.7\%~other). We use SQuAD~v1.1, not the~v2.0 superset \cite{rajpurkar:2018}, which contains unanswerable questions, since we do not expect clickbait to be ``unspoilable''.
\Nii
TriviaQA \cite{joshi:2017} contains 95,000 question--answer pairs, mostly dealing with trivia questions that are supposed to be particularly difficult to answer. These are comparable to clickbait in that many of them address rather trivial things (see Figure~\ref{twitter-clickbait-examples}).

The question answering models used in our experiments are
ALBERT \cite{lan:2019},
AllenAI-Document-QA \cite{clark:2018},
BERT (cased/uncased) \cite{devlin:2019},
Big Bird \cite{zaheer:2020},
DeBERTa (large) \cite{he:2020},
ELECTRA \cite{clark:2020},
FunnelTransformer \cite{dai:2020},
MPNet \cite{song:2020}, and
RoBERTa (base/large) \cite{liu:2019}.
Many of them are or were state of the art on the above benchmarks and implement various different architectural paradigms.

\subsection{Passage Retrieval}

Passage retrieval relaxes the question answering task a bit in the sense of allowing longer passages of text as answers (e.g., one or more sentences), rather than exact phrases or statements. Neural retrieval models, as surveyed by \citet{guo:2020} and \citet{lin:2020}, have been successfully applied to passage retrieval. One of the most important passage retrieval benchmarks is part of MS~MARCO, a series of challenges whose first edition was a large question answering task \cite{nguyen:2016}. A passage retrieval dataset of 8.8~million passages was derived for the underlying set of 100,000~questions originally submitted to Bing. This dataset formed the basis for two consecutive shared tasks at the TREC~2019 and~2020 Deep Learning tracks \cite{craswell:2019,craswell:2020}.

The passage retrieval models used in our experiments are MonoBERT \cite{nogueira:2019a,nogueira:2019b} and MonoT5 \cite{nogueira:2020} (both topped the MS~MARCO passage retrieval leaderboard once), and the classic baseline models BM25 \cite{robertson:2009} and Query Likelihood \cite{ponte:1998}, implemented in Anserini \cite{yang:2017}.

%% file: acl22-clickbait-spoiling-part3.tex
\section{Webis Clickbait Spoiling Corpus~2022}
\label{data}

To tackle clickbait spoiling for the first time, we created the Webis Clickbait Spoiling Corpus~2022 (Webis-Clickbait-22), a collection of 5,000~clickbait posts and their associated spoilers.

\subsection{Corpus Construction}

Our corpus is primarily based on five social media accounts on Twitter, Reddit, and Facebook that manually spoil clickbait: \href{https://www.reddit.com/r/savedyouaclick/}{r/savedyouaclick}, \href{https://twitter.com/huffpospoilers}{@HuffPoSpoilers}, \href{https://twitter.com/savedyouaclick}{@SavedYouAClick}, \href{https://twitter.com/upworthyspoiler}{@UpworthySpoiler}, and \href{https://www.facebook.com/StopClickBaitOfficial/}{@StopClickBaitOfficial}. With the goal of collecting 5,000~``spoilable'' clickbait posts at an expected rejection rate of around~10\% of unusable posts, 5,555~were initially collected from the accounts. Each of them was manually reviewed, and those that turned out not to be spoiled clickbait were removed (e.g., funny posts not intended to be spoilers, or posts with unavailable linked documents). The rejection rate was higher than expected, and only 4,204~posts remained.

To reach our goal of 5,000~posts, we then sampled from the Webis-Clickbait-17 corpus used in the Clickbait Challenge~2017~\cite{potthast:2018w}. The corpus contains 38,517~tweets, each of which was rated by 5~annotators on a 4-point Likert scale for clickbaitiness: ``no clickbait,'' ``slight clickbait,'' ``considerable clickbait,'' and ``heavy clickbait.'' Of the tweets, 1,845~scored an average of~0.8 or higher and can safely be considered clickbait. We selected tweets from this subset and manually spoiled them based on the linked document until our target size of 5,000~posts was reached.

\input{table-corpus-characteristics}

Thus, our final corpus consists of 4,204~posts from Twitter, Reddit, and Facebook that were spoiled by a third party specializing in this task, and 796~tweets from the Webis-Clickbait-17 corpus with an average clickbaitiness of at least~0.8 that we spoiled ourselves. For each of the 5,000~clickbait posts, we also reviewed and corrected erroneous spoilers and labeled their exact positions in the linked documents. Our internal guidelines dictated that a spoiler should be as short as possible (i.e., if one word is enough, not a whole sentence should be chosen). Since the underlying annotation task is simple, one main annotator was sufficient. Nevertheless, randomly selected as well as ambiguous cases were discussed with two additional experts among the co-authors. No systematic errors or unforeseen difficulties in solving the annotation task were identified during these discussions.

During our annotation, we found that none of the common approaches to main content extraction worked reliably for all the documents linked in the clickbait posts. Yet, clean content is a prerequisite for research on clickbait spoiling to eliminate as many confounding variables as possible. To ensure a clean corpus, one annotator manually extracted the main content of the linked documents, removing (inline) advertisements, links to related articles (e.g., ``READ ALSO:~[\ldots]'' or ``Also from CNBC~[\ldots]''), credits (e.g., ``Image credit:~[\ldots]'' or ``Photo by~[\ldots]''), and social media links (e.g., ``Subscribe to~[\ldots]'' or ``Follow us on~[\ldots]''). A random selection was reviewed to ensure high quality.

Moreover, during spoiler annotation, it turned out that there are basically three types of spoilers:
\Ni
\emph{phrase spoilers} consisting of a single word or phrase from the linked document (e.g., the first two spoilers in Figure~\ref{twitter-clickbait-examples}, but often named entity spoilers as well),
\Nii
\emph{passage spoilers} consisting of one or a few sentences of the linked document (e.g., the third spoiler in Figure~\ref{twitter-clickbait-examples}), and
\Niii
\emph{multipart spoilers} consisting of more than one non-consecutive phrases or passages of the linked document (e.g., the fourth spoiler in Figure~\ref{twitter-clickbait-examples}).
Spoiler types were also annotated by the main annotator, and randomly checked by the other two.

In sum, each of the 5,000~posts in our corpus consists of a unique~ID, the platform from which it was taken, the respective platform's post~ID, the post's text (i.e., the ``clickbait''), the URL to the linked document, the manually extracted title and paragraph-divided main content of the linked document, the manually optimized spoiler, the spoiler's character position in the main content, and the type of spoiler (phrase, passage, or multipart). In total, the annotation took about 560~hours, which marked the limit of our budget dedicated for this step.

\subsection{Corpus Statistics}

Table~\ref{table-corpus-characteristics} summarizes the main statistics of our corpus. Most spoiled clickbait posts come from Twitter~(47.5\%) and Reddit~(36\%), whereas the Facebook account contributes less~(16.5\%). Most spoilers are phrases~(42.5\%) and passages~(40\%). That there are fewer multi-part spoilers could be due to the fact that spoiler account operators prefer to spoil ``simpler'' clickbait posts. For the corpus, we also provide a fixed random 80/20/20~train/validation/test split to ensure future reproducibility and comparability with our results.

%% file: table-corpus-characteristics.tex
\begin{table*}
\small
\centering
\renewcommand{\tabcolsep}{6.25pt}
\caption{Key statistics of the Webis Clickbait Spoiling Corpus~2022 (Webis-Clickbait-22).}
\label{table-corpus-characteristics}
\begin{tabular}{@{}llrr@{\hspace{1em}}ccc@{\hspace{1em}}rrr@{\hspace{1em}}lr@{}}
\toprule
\textbf{Source} & \textbf{Spoiler} & \multicolumn{2}{@{}c@{\hspace{1em}}}{\textbf{Entries}} & \multicolumn{3}{c@{\hspace{1em}}}{\textbf{Average text length {\tiny \color{gray} $\mathbf{\pm}$ Std.Dev.}}} & \multicolumn{3}{c@{\hspace{1em}}}{\textbf{Corpus splits}} & \multicolumn{2}{c@{\hspace{1em}}}{\textbf{Top source}}\\

\cmidrule(r{1em}){5-7} \cmidrule(r{1em}){8-10} \cmidrule{11-12}

& &  &  & Post & Document & Spoiler & Train & Val. & Test & Name & Count\\

\midrule

& Phrase & 342 &  & 13.4 {\tiny \color{gray} $\mathbf{\pm}$3.6} & 433.7 {\tiny \color{gray} $\mathbf{\pm}$347.9\phantom{0}} & \phantom{0}3.0 {\tiny \color{gray} $\mathbf{\pm}$1.6\phantom{0}} & 221 & 45 & 76 & Stop Clickbait & 342\\
Facebook & Passage & 388 & & 13.4 {\tiny \color{gray} $\mathbf{\pm}$4.0} & 490.9 {\tiny \color{gray} $\mathbf{\pm}$351.5\phantom{0}} & 24.9 {\tiny \color{gray} $\mathbf{\pm}$20.0} & 231 & 73 & 84 & Stop Clickbait & 388\\
& Multipart & 94 &  & 14.2 {\tiny \color{gray} $\mathbf{\pm}$4.1} & 651.8 {\tiny \color{gray} $\mathbf{\pm}$545.2\phantom{0}} & 28.5 {\tiny \color{gray} $\mathbf{\pm}$33.0} & 68 & 12 & 14 & Stop Clickbait & 94\\

\midrule
& Phrase & 688 &  & 13.2 {\tiny \color{gray} $\mathbf{\pm}$4.0} & 584.6 {\tiny \color{gray} $\mathbf{\pm}$798.6\phantom{0}} & \phantom{0}2.8 {\tiny \color{gray} $\mathbf{\pm}$1.6\phantom{0}} & 455 & 109 & 124 & savedyouaclick & 688\\
Reddit & Passage & 859 & & 13.1 {\tiny \color{gray} $\mathbf{\pm}$4.0} & 657.2 {\tiny \color{gray} $\mathbf{\pm}$1004.7} & 25.4 {\tiny \color{gray} $\mathbf{\pm}$20.3} & 533 & 148 & 178 & savedyouaclick & 859\\
& Multipart & 250 &  & 12.8 {\tiny \color{gray} $\mathbf{\pm}$4.4} & 991.7 {\tiny \color{gray} $\mathbf{\pm}$899.5\phantom{0}} & 32.7 {\tiny \color{gray} $\mathbf{\pm}$36.2} & 162 & 46 & 42 & savedyouaclick & 250\\

\midrule
& Phrase & 1,095 &  & 11.0 {\tiny \color{gray} $\mathbf{\pm}$3.4} & 479.1 {\tiny \color{gray} $\mathbf{\pm}$502.9\phantom{0}} & \phantom{0}2.7 {\tiny \color{gray} $\mathbf{\pm}$1.7\phantom{0}} & 691 & 181 & 223 & HuffPoSpoilers & 794\\
Twitter & Passage & 752 &  & 10.3 {\tiny \color{gray} $\mathbf{\pm}$4.2} & 597.4 {\tiny \color{gray} $\mathbf{\pm}$605.8\phantom{0}} & 22.3 {\tiny \color{gray} $\mathbf{\pm}$ 13.5} & 510 & 101 & 141 & HuffPoSpoilers & 328\\
& Multipart & 532 &  & 11.5 {\tiny \color{gray} $\mathbf{\pm}$3.8} & 884.0 {\tiny \color{gray} $\mathbf{\pm}$930.3\phantom{0}} & 35.4 {\tiny \color{gray} $\mathbf{\pm}$34.4} & 329 & 85 & 118 & HuffPoSpoilers & 148\\

\midrule
& Phrase & 2,125 &  & 12.1 {\tiny \color{gray} $\mathbf{\pm}$3.8} & 505.9 {\tiny \color{gray} $\mathbf{\pm}$599.4} & 2.8 {\tiny \color{gray} $\mathbf{\pm}$1.6} & 1,367 & 335 & 423 & HuffPoSpoilers & 794\\
$\sum$ & Passage & 1,999 &  & 12.1 {\tiny \color{gray} $\mathbf{\pm}$4.3} & 602.4 {\tiny \color{gray} $\mathbf{\pm}$774.0} & 24.1 {\tiny \color{gray} $\mathbf{\pm}$18.1} & 1,274 & 322 & 403 & savedyouaclick & 859\\
& Multipart & 876 &  & 12.2 {\tiny \color{gray} $\mathbf{\pm}$4.1} & 889.8 {\tiny \color{gray} $\mathbf{\pm}$892.2} & 33.9 {\tiny \color{gray} $\mathbf{\pm}$34.8} & 559 & 143 & 174 & savedyouaclick & 250\\

\bottomrule
\end{tabular}
\end{table*}

%% file: acl22-clickbait-spoiling-part4.tex
\section{Type-dependent Clickbait Spoiling}
\label{approach}

Our approach to clickbait spoiling is based on the observation that there are three types of spoilers:
\Ni
phrase spoilers,
\Nii
passage spoilers, and
\Niii
multipart spoilers.
We assume that different tailored approaches will work best for each spoiler type. However, an important prerequisite for this is the corresponding classification of clickbait. Therefore, we first investigate how well the spoiler type of a clickbait post can be predicted (Section~\ref{spoiler-type-classification}).

The generation of phrase and passage spoilers for a given clickbait post is similar in that the solution to the problem in both cases amounts to extracting a coherent piece of text from the linked document. To this end, there are a variety of existing approaches in related disciplines whose output is either a phrase or a passage, and which may be adapted to clickbait spoiling. We therefore investigate whether phrase spoilers can be identified by conventional question answering methods (i.e., we treat a clickbait post as a ``question'' to which a phrase of the linked document should be returned as the ``answer''; Section~\ref{phrase-spoiling}), and whether passage spoilers can be identified by conventional passage retrieval methods (i.e., we treat a clickbait post as a ``query'' and the paragraphs of the linked document as the collection from which to retrieve the best ``passage''; Section~\ref{passage-spoiling}). In our evaluation, we focus on phrase and passage spoilers and also examine the abilities of the above question answering and passage retrieval methods to serve as one-size-fits-all solutions for phrases and passages. For multipart spoilers, a novel approach will be needed, which is beyond the scope of our current work but an interesting direction for the future.

\subsection{Spoiler Type Classification}
\label{spoiler-type-classification}

For the spoiler classification subtask, we experimented with classic feature-based models (Na{\"i}ve Bayes, Logistic Regression, SVM) and the neural models BERT-, DeBERTa-, and RoBERTa.

As feature types for the classic models, we use \tf- and \tfidf-weighted word and POS tag uni- and bigrams from the clickbait post and \tfidf-weighted word and POS tag uni- and bigrams from the linked document. We include features from the linked document, since it has to be analyzed for the spoiler generation anyway. The \idf values are calculated on the OpenWebText~corpus \cite{gokaslan:2019} to prevent any bias from the comparatively small size of our corpus. 

The input for the neural models is a post concatenated with the main content of the linked document.

\subsection{Phrase Spoiler Generation}
\label{phrase-spoiling}

Viewing a clickbait post for which a phrase spoiler should be derived as a ``question'' and the linked document as potentially containing an ``answer'', phrase spoiler generation can be tackled by question answering methods. We therefore employ ten state-of-the-art question answering methods trained on the SQuAD~data and fine-tune them on our new clickbait spoiling training set: ALBERT, BERT (cased/uncased), BigBird, DeBERTa (large), ELECTRA, FunnelTransformer, MPNet, and RoBERTa (base/large).

\subsection{Passage Spoiler Generation}
\label{passage-spoiling}

Treating the clickbait post whose spoiler type is a passage as a ``query'' for which the ``most relevant'' passage from the linked document is to be retrieved, passage spoiler generation can be tackled by passage retrieval methods. We therefore use ten state-of-the-art passage retrieval approaches trained on the MS~MARCO~data: BM25 and QLD in four variants each (alone or with RM3/Ax/PRF query expansion), MonoBERT, and MonoT5. In addition, we also adapt all of the above question answering models to retrieve passages by simply considering the passage as the returned result from which the question answering model extracts its answer.

%% file: acl22-clickbait-spoiling-part5.tex
\section{\kern-0.25em\fontsize{11.8pt}{13pt}\selectfont Evaluation of Spoiler Type Classification}
\label{evaluation-of-spoiler-type-generation}

In our evaluation, we assume a setup in which a previous clickbait detection would have (perfectly) identified posts as clickbait. To then evaluate the effectiveness of spoiler type classification on such detected clickbait posts, we conduct three experiments:
\Ni
multi-class,
\Nii
one-vs-rest, and
\Niii
one-vs-one for the types of phrase and passage spoilers.

\input{table-classification-effectiveness-all-pairs}

In all cases, the hyperparameters of the six studied classifiers were optimized based on the validation set of our corpus. For the three feature-based approaches, a chi-square feature selection step selected all post-based features and 70\% of the document-based features. The post-based features are weighted 4-times higher than the document-based features. Most hyperparameters of the transformer models were left at their default values, but a grid search was used to find the most effective combination of learning rate (1e-5, 4e-5, 1e-4), warm-up ratio (0.02, 0.06, and 0.1), stack size (8, 16, and~32), number of epochs (1 to~10), and maximum sequence length (256, 384, 512).

Table~\ref{table-classification-effectiveness-all-pairs} shows the balanced accuracy of the six classifiers. All are less effective in the multi-class setting than in the one-vs-rest settings and the transformer-based classifiers are clearly more effective than the feature-based ones; DeBERTa is best in the multi-class setting (accuracy of~73.63) and RoBERTa in the one-vs-rest ones (79.12~to~80.39).

\input{table-classification-effectiveness}

Table~\ref{table-classification-effectiveness} shows the accuracy of the six classifiers on the 826~test posts with phrase and passage spoilers (almost balanced setup, since there is hardly any class imbalance). Again, the transformer-based classifiers clearly are more effective than the feature-based ones; with RoBERTa achieving the best accuracy of~80.39.

The substantial improvements of DeBERTa and RoBERTa over the feature-based classifiers in all settings (about 9--10~accuracy points) indicates that classifying the clickbait spoiler type requires more advanced language ``understanding'' than what is encoded in the basic features that the Na{\"i}ve Bayes, SVM, or logistic regression classifiers used.

%% file: table-classification-effectiveness-all-pairs.tex
\begin{table}
\small
\centering
\renewcommand{\tabcolsep}{8pt}
\caption{Effectiveness of spoiler type classification in the multi-class (first column) and one-vs-rest settings on 1000 test posts (training:~3200; validation:~800).}
\label{table-classification-effectiveness-all-pairs}
\begin{tabular}{@{}l@{\quad}lcccc@{}}
\toprule
\bfseries Model & \multicolumn{5}{@{}c@{}}{\textbf{Balanced accuracy} (0, 1, 2 indicate class labels)}\\
\cmidrule{2-6}
& Phrase    & 0 & 1 & 0 & 0 \\
& Passage   & 1 & 0 & 1 & 0 \\
& Multipart & 2 & 0 & 0 & 1 \\

\midrule

\multicolumn{2}{@{}l}{Na{\"i}ve Bayes} & 56.15 & 65.03 & 62.50 & 64.82\\
\multicolumn{2}{@{}l}{SVM} & 59.62 & 68.03 & 68.70 & 70.28\\
\multicolumn{2}{@{}l}{Log. Regression} & 60.04 & 68.04 & 69.33 & 71.26\\

\midrule

\multicolumn{2}{@{}l}{BERT} & 67.84 & 74.06 & 75.70 & 75.56\\
\multicolumn{2}{@{}l}{DeBERTa}  & \bfseries 73.63 & 78.39 & 78.65 & 77.93\\
\multicolumn{2}{@{}l}{RoBERTa}  & 71.57 & \bfseries 80.39 & \bfseries 79.30 & \bfseries 79.12\\

\bottomrule
\end{tabular}
\end{table}

%% file: table-classification-effectiveness.tex
\begin{table}
\small
\centering
\renewcommand{\tabcolsep}{8pt}
\caption{Effectiveness of spoiler type classification in the one-vs-one (phrase-vs-passage) setting on 826~test posts (training: 2,641; validation: 657).}
\label{table-classification-effectiveness}
\begin{tabular}{@{}l@{\qquad}cccc@{\qquad}c@{}}
\toprule
\bfseries Model & \multicolumn{5}{@{}c@{}}{\textbf{Effectiveness}}\\
\cmidrule{2-6}
& TP & TN & FP & FN & Acc.\\
\midrule
Na{\"i}ve Bayes & 298 & 256 &           147 &           125 &           67.07 \\
SVM             & 311 & 264 &           139 &           112 &           69.61 \\
Log. Regression & 306 & 273 &           130 &           117 &           70.10 \\
\midrule
BERT            & 315 & 315 & \phantom{0}88 &           108 &           76.27 \\
DeBERTa         & 318 & 335 & \phantom{0}68 &           105 &           79.06 \\
RoBERTa         & 332 & 332 & \phantom{0}71 & \phantom{0}91 & \bfseries 80.39 \\
\bottomrule
\end{tabular}
\end{table}

%% file: acl22-clickbait-spoiling-part6.tex
\section{Evaluation of Spoiler Generation}
\label{evaluation-of-spoiler-generation}

To assess the effectiveness of the question answering and passage retrieval methods for clickbait spoiling, we evaluate both for their respective intended spoiler types, but each also for the respective other spoiler type. Multipart spoilers are deferred to future work. We continue to assume that prior clickbait detection (perfectly) identifies clickbait posts as such. Our evaluation of the generated spoilers includes quantitative and qualitative assessments (Section~\ref{assessment-metrics}). In a pilot study with ten question answering and ten passage retrieval models at their default settings, two models in each category dominate the respective others (Section~\ref{pilot-study}). The computationally expensive step of hyperparameter optimization is restricted to these four models plus two baselines (Section~\ref{model-tuning-and-selection}). Then, the effectiveness of spoiling clickbait posts dependent on spoiler type is evaluated (Sections~\ref{evaluation-phrase-spoilers} and~\ref{evaluation-passage-spoilers}), and compared to an end-to-end clickbait spoiling setup independent of spoiler type (section~\ref{end-to-end-system}).

\subsection{Quantitative and Qualitative Assessment}
\label{assessment-metrics}

We introduce the measures used to evaluate generated spoilers and how we manually determined thresholds for them above which a generated spoiler is considered as ``correct''.

\paragraph{Evaluation measures.}
To assess the quantitative correspondence between a derived spoiler and the ground truth, we use three question answering-oriented and one passage retrieval-oriented measure: BLEU-4 \cite{papineni:2002}, METEOR \cite{banerjee:2005} in its extended version of \citet{denkowski:2014}, BERTScore \cite{zhang:2020}, and Precision@1.

The three question answering-oriented measures each calculate a (penalized) harmonic mean of measure-specific definitions of precision and recall when comparing a generated spoiler to the ground truth. In case of BLEU-4, the overlap of word 1-~to 4-grams is determined (if the length~$n$ of a generated spoiler is less than 4~words, we compute BLEU-$n$), in case of METEOR the overlap of word 1-grams, and in case of BERTScore the best matching embeddings of word pairs. Note that in their original formulation, BLEU-4 and METEOR penalize the score, the more the $n$-gram order differs. To arrange the measures on a spectrum from calculating predominantly syntactic (BLEU-4) to predominantly semantic similarity (BERTScore), we omit METEOR's penalization term.

The question answering-oriented measures are not really suited to assess the effectiveness of passage retrieval models since a retrieved passage is often longer than the ground truth spoiler. Therefore, we also use Precision@1 to measure whether the top-ranked passage contains the ground truth spoiler (all phrase spoilers and 98\%~of the passage spoilers come from a single passage; for the other passage spoilers, we consider all containing passages as relevant). To calculate the Precision@1 of question answering models, we use the first passage that contains the returned spoiler.

\input{table-high-confidence-thresholds}
\input{table-spoiling-effectiveness-pilot-study}

\paragraph{High-confidence thresholds.}
Candidates with higher scores on the question answering-oriented measures BLEU-4, METEOR, and BERTScore are closer to the ground truth. However, it is unclear what score threshold a particular spoiler candidate has to exceed so that it would be considered a true positive in a manual analysis. Determining such thresholds enables ``high confidence'' estimations of how many correct spoilers an approach generates without having to manually check its outputs each time with each new variant.

In a pilot study, we thus determined such thresholds by running all question answering models (cf.\ Section~\ref{phrase-spoiling} and~\ref{passage-spoiling}) on a random sample of 500~clickbait posts with phrase spoilers and 500~with passage spoilers. For each post, a random spoiler generated by a question answering model and a random spoiler generated by a passage retrieval model were manually checked for whether they could be viewed as correct. Table~\ref{table-high-confidence-thresholds} shows the number of manually determined false positives and false negatives for different thresholds of BLEU-4, METEOR, and BERTScore. The manually selected subjective thresholds (FP/FN in bold) for each combination of measure, spoiler type, and model type (question answering or passage retrieval) minimize the false positives at a rate where being more strict would incur too many false negatives. For instance, for phrase spoilers and BLEU-4, we set the question answering model threshold at~50\% since a more strict threshold of~60\% does not reduce the false positives but increases the false negatives.

In addition to reporting quantitative mean effectiveness scores, applying the determined thresholds helps to estimate how many of the spoilers of a model would be perceived as ``good'' by human readers. This corresponds to a conservative assessment, since we believe that a model should only be deployed to production if it has been tuned to not return a spoiler if in doubt about its correctness; also probably somewhat minimizing the otherwise possible spread of auto-generated misinformation.

\subsection{Pilot Study for Model Selection}
\label{pilot-study}

In a pilot study on 1,000~clickbait posts (800~training, 200~validation), we compare ten question answering and ten passage retrieval models (cf.\ Table~\ref{table-spoiling-effectiveness-pilot-study}) at their default settings to select models for subsequent experiments with more extensive (and expensive) hyperparameter tuning. The question answering models were or are among the most effective in the SQuAD and TriviaQA question answering benchmarks. In our setup, they return a piece of text from the linked document as an ``answer'' to the clickbait post as the ``query''. As passage retrieval models, we empoly MonoBERT and MonoT5 using their PyGaggle%
\footnote{\url{https://github.com/castorini/pygaggle}}
implementations, and eight variants of the popular baseline retrieval models~BM25 and~QLD using their Anserini implementations~\cite{yang:2017}. These models return the most ``relevant'' paragraph from the linked document for the clickbait post as the ``query''.

\input{table-spoiling-effectiveness}

Using Nvidia A100~GPUs, the question answering models were first fine-tuned on SQuAD~v1.1 and then on the pilot training data. This was the most effective setup from an ablation study with other fine-tuning regimes (e.g., the phrase spoiler BERTScore for RoBERTa-large dropped from~84.04 to~69.91 when only fine-tuned on our pilot study data, to~64.61 when only fine-tuned on SQuAD, and to~46.60 without fine-tuning). Interestingly, the models' SQuAD~effectiveness does not predict their spoiling effectiveness (e.g., RoBERTa-base and FunnelTransformer were tied on SQuAD, but RoBERTa-base is more effective at spoiling). This indicates the importance of the pilot study.

Table~\ref{table-spoiling-effectiveness-pilot-study} shows the pilot study effectiveness of all models on the 200~validation posts. RoBERTa-large (for phrasal spoilers) and DeBERTa-large (for passage spoilers) are the most effective. Among the passage retrieval models, MonoBERT and MonoT5 achieve the best scores. Contrary to our original assumption that passage retrieval models might be particularly well-suited to identify passage spoilers, MonoBERT and MonoT5 have similar Precision@1 scores on both phrase and passage spoilers and are substantially less effective than the best question answering models (e.g., DeBERTa-large has a Precision@1 of~48.39 for passage spoilers compared to~31.18 for MonoBERT).

\subsection{Tuning the Selected Models}
\label{model-tuning-and-selection}

Given the pilot study results, six models are selected for a more extensive hyperparameter tuning: the best two question answering models (DeBERTa-large was best for phrase spoilers, RoBERTa-large for passage spoilers) plus BERT as baseline, as well as the best two passage retrieval models (MonoBERT and MonoT5) plus BM25 as baseline.

As the ablation study in our pilot study showed that fine-tuning the question answering models on SQuAD first and then on our corpus works best, we apply this fine-tuning regime to DeBERTa-large, RoBERTa-large, and BERT using the clickbait spoiling training data (depending on the experiment, either only the phrase spoilers, only the passage spoilers, or both combined). Most hyperparameters of DeBERTa-large, RoBERTa-large, BERT, MonoBERT, and MonoT5 are left at their defaults, but a grid search is run to find the most effective combination of learning rate (1e-5, 4e-5, 1e-4), warmup ratio (0.02, 0.06, 0.1), batch size (8, 16, 32), number of epochs (1 to~10), and maximum sequence length (256, 384, 512). For BM25, we try combinations of~$k_1$ from~0.1 to~0.4 and~$b$ from~0.1 to~1.0 with a step size of~0.1.

\subsection{Effectiveness on Phrase Spoilers}
\label{evaluation-phrase-spoilers}

The `Phrase Spoilers' column group in Table~\ref{table-spoiling-effectiveness} shows the effectiveness of the selected question answering and passage retrieval models on the 423~test clickbait posts with phrase spoilers. Given the ground-truth spoiler, we report the predicted spoilers' average BLEU-4, METEOR, BERTScore, and Precision@1 (using 1,367~posts with phrase spoilers for training and 335~posts for validation to tune the hyperparameters; cf.\ Table~\ref{table-corpus-characteristics}). 

Overall, DeBERTa-large is the most effective model for phrase spoilers. Based on our high-confidence score thresholds, it generates the correct spoiler for~250--300 of the 423~test posts (i.e., for about 60--70\% of the cases) according to a BERTScore or BLEU-4 evaluation. Similar to our pilot study, the passage retrieval models are comparably ineffective in identifying phrase spoilers. Among them, MonoT5 achieves the highest scores but is even substantially less effective than the question answering baseline~BERT. For instance, with a BLEU-4 of~58.89 and probably 257~correct spoilers (61\% of the 423~test posts), BERT is way ahead of MonoT5 with a BLEU-4 of~4.95 and only 82~probably correct spoilers (19\% of the 423 posts).

\subsection{Effectiveness on Passage Spoilers}
\label{evaluation-passage-spoilers}

The `Passage Spoilers' column group in Table~\ref{table-spoiling-effectiveness} shows the effectiveness of the selected passage retrieval models on the 403~test clickbait posts with passage spoilers (using 1,274~and 322~posts for training and validation). The numbers of probably correct spoilers are lower for all models compared to the phrase spoilers (even the higher amount of probably correct passage spoilers of the passage retrieval models according to their BERTScore threshold are still worse than the estimated probably correct phrase spoilers according to BLEU-4 or METEOR). Similar to the pilot study, all question answering models are also substantially more effective on passage spoilers than the passage retrieval models. Overall, DeBERTa-large and RoBERTa-large achieve the highest Precision@1~scores and the highest amount of probably correct passage spoilers (about 35--41\% of the passage spoilers are correctly identified according to our high-confidence thresholds).

\subsection{Effectiveness of the End-to-End System}
\label{end-to-end-system}
\input{table-end-to-end-effectiveness}

We evaluate the entire spoiling pipeline using all 826~phrase and passage test posts by comparing two-step pipelines that first classify the spoiler type to then select an appropriately trained spoiler model (trained on the respective type) and single-step approaches that skip the spoiler type classification and simply run the same spoiler model on all posts (trained on the complete training data). For the two-step pipelines, we experiment with two variants: \Ni using an artificial classifier that returns perfect oracle-style answers about a post's type, and \Nii using the best RoBERTa-based phrase-vs-passage classifier from Section~\ref{evaluation-of-spoiler-type-generation}.

Since the passage retrieval models were less effective in our spoiler experiments (cf.\ Table~\ref{table-spoiling-effectiveness}), we report results only for pipelines with question answering models. In the two-step pipelines the respective question answering models are fine-tuned on the respective spoiler types, in the single-step approach on the combined training data.

Table~\ref{table-end-to-end-effectiveness} shows the achieved end-to-end effectiveness values. The individual two-step pipelines with oracle type classification (row group `Oracle') are substantially more effective than their single-step counterparts without type classification (row group `None') that again are more effective than the respective two-step pipelines with ``real'' RoBERTa-based type classification (row group `Classif.'). Overall, the DeBERTa pipeline with oracle classifier achieves an estimated amount of about 50--55\% correctly spoiled posts~(i.e., 411~to 457 of~826). This result confirms that classifying the required spoiler type can be beneficial for clickbait spoiling. Still, among the currently realistically applicable end-to-end spoiling approaches (with RoBERTa type classification or without spoiler type classification), the one-step DeBERTa approach without spoiler type classification is the most effective according to the number of probably correctly spoiled posts (382~to 409~of the 826~posts, i.e., 46--50\%). This indicates that the currently best RoBERTa-based spoiler type classifier with its accuracy of~80.39\% is still not good enough to result in an end-to-end system that actually benefits from spoiler type classification.

Our results show that effectively spoiling clickbait with question answering models is possible in practice but also that there is still room for improvements (e.g., improved spoiler type classification, improved spoiler generation for the individual types, and taking multipart spoilers into account).

%% file: table-high-confidence-thresholds.tex
\begin{table}
\small
\centering
\renewcommand{\tabcolsep}{1.9pt}
\caption{Manually determined numbers of false positives/negatives~(FP/FN) on 500~sampled clickbait posts with phrase spoilers and 500~with passage spoilers for question answering (top row group) and passage retrieval models (bottom row group), dependent on score threshold (Thresh.), spoiler type, and effectiveness measure (BL4 = BLEU-4, MET = METEOR, BSc.\ = BERTScore). The thresholds selected for subsequent assessment are indicated by bold FP/FN numbers.}
\label{table-high-confidence-thresholds}
\begin{tabular}{@{}l@{\kern-1em}rrrrrr@{\hspace*{1em}}rrrrrr@{}}
\toprule
\bfseries Thresh. & \multicolumn{6}{@{}c@{\hspace*{1em}}}{\bfseries Phrase Spoilers} & \multicolumn{6}{@{}c@{}}{\bfseries Passage Spoilers} \\
\cmidrule(r@{1em}){2-7}\cmidrule{8-13}
& \multicolumn{2}{@{}c}{BL4} & \multicolumn{2}{@{}c}{MET} & \multicolumn{2}{@{}c@{\hspace*{1em}}}{BSc.} & \multicolumn{2}{@{}c}{BL4} & \multicolumn{2}{@{}c}{MET} & \multicolumn{2}{@{}c@{}}{BSc.} \\
\cmidrule(r@{\tabcolsep}){2-3}\cmidrule(l@{\tabcolsep}r@{\tabcolsep}){4-5}\cmidrule(l@{\tabcolsep}r@{1em}){6-7}\cmidrule(r@{\tabcolsep}){8-9}\cmidrule(l@{\tabcolsep}r@{\tabcolsep}){10-11}\cmidrule(l@{\tabcolsep}){12-13}
     & FP & FN & FP & FN & \multicolumn{1}{c}{FP} & FN & FP & FN & \multicolumn{1}{c}{FP} & FN & \multicolumn{1}{c}{FP} & FN \\
\midrule
10\% & 11 & 11 & 18 &  7 & 238 &  0 & 5 & 44 & 168 & 15 & 399 &  0 \\
20\% &  7 & 14 & 16 &  7 & 234 &  0 & 3 & 48 &  67 & 27 & 325 &  3 \\
30\% &  7 & 14 & 14 &  9 & 165 &  1 & \bfseries 1 & \bfseries 51 &  31 & 35 & 134 & 21 \\
40\% &  2 & 27 &  8 & 13 &  59 &  6 & 0 & 55 &  15 & 39 &  18 & 38 \\
50\% & \bfseries 2 & \bfseries 27 &  2 & 28 &  24 & 14 & 0 & 60 &   9 & 42 &   5 & 51 \\
60\% &  2 & 30 &  3 & 31 &  11 & 25 & 0 & 64 &   4 & 57 & \bfseries 1 & \bfseries 59 \\
70\% &  1 & 33 & \bfseries 2 & \bfseries 31 &   6 & 36 & 0 & 66 & \bfseries 1 & \bfseries 54 &   0 & 66 \\
80\% &  1 & 34 &  0 & 37 & \bfseries 1 & \bfseries 40 & 0 & 66 &   0 & 61 &   0 & 73 \\
\midrule
\phantom{0}5\% & 8 &  40 & 28 &  64 & 208 &   0 &  \bfseries 0 &  \bfseries 95 & 225 &  10 & 355 &  0 \\
10\% & \bfseries 4 & \bfseries 104 &  \bfseries 8 & \bfseries 108 & 180 &  60 &  0 &  95 & 140 &  30 & 355 &  0 \\
20\% & 0 & 184 &  0 & 164 &  44 & 144 &  0 &  95 &  35 &  65 & 305 &  15 \\
30\% & 0 & 188 &  0 & 184 &   \bfseries 0 & \bfseries 176 &  0 & 105 &   \bfseries 5 &  \bfseries 90 & 145 &  55 \\
40\% & 0 & 188 &  0 & 188 &   0 & 188 &  0 & 115 &   5 & 105 &  20 &  95 \\
50\% & 0 & 192 &  0 & 188 &   0 & 192 &  0 & 120 &   5 & 110 &   \bfseries 5 & \bfseries 105 \\
60\% & 0 & 192 &  0 & 192 &   0 & 192 &  0 & 125 &   0 & 120 &   5 & 130 \\
\bottomrule
\end{tabular}
\end{table}

%% file: table-spoiling-effectiveness-pilot-study.tex
\begin{table*}
\small
\centering
\renewcommand{\tabcolsep}{4.5pt}
\caption{Pilot study spoiling effectiveness of question answering and passage retrieval models on 200~validation posts (models ordered lexicographically). The bracketed numbers indicate the expected number of true positives as per our pre-determined high-confidence score thresholds; P@1 is the Precision@1. The models DeBERTa-large and RoBERTa-large, as well as MonoBERT and MonoT5 are the most effective in their groups.}
\label{table-spoiling-effectiveness-pilot-study}
\begin{tabular}{@{}llrrrrrrrr@{}}
\toprule
\bfseries Type & \bfseries Model & \multicolumn{4}{@{}c@{\quad}}{\textbf{Phrase Spoilers} $\;\,$($n$ = 97)} & \multicolumn{4}{@{}c@{\quad}}{\textbf{Passage Spoilers} $\;\,$($n$ = 103)} \\

\cmidrule(l{\tabcolsep}r{\tabcolsep}){3-6}\cmidrule(l{\tabcolsep}){7-10}

&& BLEU-4 & METEOR & BERTScore & P@1 & BLEU-4 & METEOR & BERTScore & P@1\\
\midrule
\multirow{10}{*}{\parbox{1.4cm}{Question\\Answering}}

& ALBERT        & 63.82           (50) & 55.97           (49) & 74.07           (46) &           63.64 & 24.51           (33) & 38.42           (27) & 44.61           (24) &  38.71 \\
& BERT-cased    & 60.27           (49) & 58.87           (47) & 73.55           (44) &           59.09 & 17.65           (22) & 28.09           (20) & 40.30           (16) &  27.96 \\
& BERT-uncased  & 62.36           (49) & 53.17           (47) & 75.87           (47) &           60.23 & 18.05           (22) & 32.50           (20) & 39.86           (18) &  32.26 \\
& Big Bird       & 69.21           (55) & 64.80           (54) & 77.39           (49) &           63.64 & 23.89           (30) & 36.20           (28) & 44.55           (27) &  43.01 \\
& DeBERTa-large & 70.19           (57) & 65.08           (56) & 78.02           (50) &           65.91 & 29.52 \bfseries (38) & 43.72 \bfseries (36) & 49.63 \bfseries (37) &  \bfseries 48.39 \\
& ELECTRA       & 69.10           (55) & 65.97           (53) & 79.26           (51) &           65.91 & 25.78           (32) & 39.87           (30) & 46.64           (27) &  43.01 \\
& Funnel-Transf.   & 68.31           (54) & 63.89           (53) & 78.78           (51) &           64.77 & 28.59           (36) & 40.95           (32) & 47.93           (29) &  40.86 \\
& MPNet         & 72.92           (58) & 65.90           (57) & 80.26           (55) &           69.32 & 30.16           (36) & 40.68           (35) & 50.07           (32) &  40.86 \\
& RoBERTa-base  & 73.02           (59) & 65.56           (57) & 80.39           (54) &           65.91 & 27.61           (35) & 41.55           (35) & 48.76           (30) &  44.09 \\
& RoBERTa-large & 79.47 \bfseries (66) & 78.61 \bfseries (61) & 84.04 \bfseries (58) & \bfseries 70.45 & 29.58           (35) & 43.49           (32) & 48.65           (32) &  44.09 \\
\midrule
\multirow{10}{*}{\parbox{1.4cm}{Passage\\ Retrieval}}

& BM25          & \phantom{0}3.49 (10) & \phantom{0}3.67 (10) & 17.73 \phantom{0}\bfseries (2) & 5.68 & 11.49 (22) & 22.64 (21) & 36.80 (12) &  9.68 \\
& BM25+Ax       & \phantom{0}3.39 (10) & \phantom{0}3.57 \phantom{0}(9) & 18.07 \phantom{0}\bfseries (2) &  5.68 & 11.27 (21) & 22.46 (19) & 36.51 (12) &  9.94\\
& BM25+PRF      & \phantom{0}3.25 (10) & \phantom{0}3.21 \phantom{0}(9) & 18.03 \phantom{0}\bfseries (2) &  5.13 & \phantom{0}9.68 (20) & 21.10 (17) & 35.44 (11)&  8.84 \\
& BM25+RM3      & \phantom{0}3.43 (10) & \phantom{0}3.62 \phantom{0}(9) & 17.14 \phantom{0}\bfseries (2) &  5.13 & 10.06 (21) & 21.03 (20) & 35.56 (11) &  8.84 \\
& MonoBERT      & \phantom{0}3.42 \bfseries(11) & \phantom{0}4.13 \bfseries (12) & 18.32  \phantom{0}(1) &  \bfseries 32.95 & 14.55 \bfseries (29) & 26.86 (25) & 38.10 (15)&  \bfseries 31.18 \\
& MonoT5        & \phantom{0}3.16 \phantom{0}(9) & \phantom{0}4.19 (11) & 18.30 \phantom{0}(0) &  31.82 & 14.27 \bfseries (29) & 26.70 \bfseries (26) & 38.94 \bfseries (17)&  29.03 \\
& QLD           & \phantom{0}2.51 \phantom{0}(7) & \phantom{0}2.69 \phantom{0}(7) & 17.24 \phantom{0}(0) &  12.50 & 10.94 (25) & 17.80 (18) & 36.70 (11) &  19.35 \\
& QLD+Ax        & \phantom{0}2.61 \phantom{0}(7) & \phantom{0}2.71 \phantom{0}(7) & 17.10 \phantom{0}(0) &  12.50 & \phantom{0}9.68 (20) & 17.84 (18) & 36.68 (11) &  8.84 \\ 
& QLD+PRF       & \phantom{0}2.60 \phantom{0}(7) & \phantom{0}2.70 \phantom{0}(7) & 17.13 \phantom{0}(0) &  11.94 & 10.86 (25) & 17.52 (18) & 36.46 (11) &  17.67 \\
& QLD+RM3       & \phantom{0}2.41 \phantom{0}(7) & \phantom{0}2.54 \phantom{0}(7) &  16.97 \phantom{0}(0) & 11.39 & 10.66 (25) & 17.54 (18) & 36.13 (11) &  17.12 \\

\bottomrule
\end{tabular}
\end{table*}

%% file: table-spoiling-effectiveness.tex
\begin{table*}
\small
\centering
\renewcommand{\tabcolsep}{3.35pt}
\caption{Effectiveness on the 826~test clickbait posts with phrase and passage spoilers. The bracketed numbers indicate the expected number of true positives as per our pre-determined high-confidence score thresholds; P@1 is the Precision@1. Overall, DeBERTa-large and RoBERTa-large are the most effective models.}
\label{table-spoiling-effectiveness}
\begin{tabular}{@{}llrrrrrrrr@{}}
\toprule
\bfseries Type & \bfseries Model & \multicolumn{4}{@{}c@{\quad}}{\textbf{Phrase Spoilers} $\;\,$($n$ = 423)} & \multicolumn{4}{@{}c@{\quad}}{\textbf{Passage Spoilers} $\;\,$($n$ = 403)} \\

\cmidrule(l{\tabcolsep}r{\tabcolsep}){3-6}\cmidrule(l{\tabcolsep}){7-10}

&& BLEU-4 & METEOR & BERTScore & P@1 & BLEU-4 & METEOR & BERTScore & P@1 \\
\midrule
\multirow{3}{*}{\parbox{1.4cm}{Question\\Answering}}
& BERT (baseline) &           58.89 (257) &           56.75 (266) &           71.06 (215) & 66.67 &            21.59 (110) &            35.49            (100) &            44.38            (109) &  42.43 \\
& DeBERTa-large   & \bfseries 68.80 (300) & \bfseries 67.93 (298) & \bfseries 77.03 (250) &  \bfseries 75.65 & \bfseries 31.44 (157) & {\bfseries 46.06}           (142) & {\bfseries 51.06}           (161) & \bfseries 54.84 \\
& RoBERTa-large   &           65.70 (290) &           66.15 (293) &           74.81 (233) & 72.58 &           29.61 (148) &            45.20  \bfseries (145) &            49.99  \bfseries (167) &  53.85 \\

\midrule

\multirow{3}{*}{\parbox{1.4cm}{Passage\\Retrieval}}
& BM25 (baseline) & \phantom{0}3.40 \phantom{0}(55) & \phantom{0}5.06 \phantom{0}(83) & 19.94 \phantom{0}(12) &  8.27 & \phantom{0}7.91 \phantom{0}(53) & 20.19 \phantom{0}(61) & 34.71 \phantom{0}(42) &  4.22 \\
& MonoBERT        & \phantom{0}4.20 \phantom{0}(72) & \phantom{0}6.12 (103)           & 20.66 \phantom{0}(11) &  42.08 &           10.43 \phantom{0}(74) & 22.37 \phantom{0}(75) & 36.58 \phantom{0}(46) &  26.05 \\
& MonoT5          & \phantom{0}4.95 \phantom{0}(82) & \phantom{0}6.47 (115)           & 20.98 \phantom{0}(16) &  43.97 &           10.58 \phantom{0}(74) & 22.02 \phantom{0}(74) & 36.70 \phantom{0}(46) &  29.03 \\
\bottomrule
\end{tabular}
\end{table*}

%% file: table-end-to-end-effectiveness.tex
\begin{table}
\centering

\renewcommand{\tabcolsep}{3.3pt}
\small
\caption{End-to-end effectiveness on the 826~phrase and passage test posts. Spoiling models that classify the spoiler type to then select an appropriately trained spoiler model (`Classif.', using the most effective spoiler type classifier), models without spoiler type classification (`None'), and unrealistic models with perfect-accuracy type classification (`Oracle').} \label{table-end-to-end-effectiveness}
\begin{tabular}{@{}l@{\hspace*{.1cm}}lcccc@{}}
\toprule
& \bfseries Model & \multicolumn{4}{c@{}}{\textbf{End-to-End Effectiveness}}\\

\cmidrule(l{\tabcolsep}){3-6}

& & BLEU-4 & METEOR & BERTScore & P@1 \\
\midrule
\multirow{3}{*}{\rotatebox[origin=c]{90}{\parbox[c]{3.3em}{\centering \textbf{Classif.}}}}
& BERT    &           35.95 (311) &           34.25 (303) &           53.86 (294) & 52.66 \\
& DeBERTa &           44.98 (392) &           44.32 (377) &           59.18 (378) & 63.44 \\
& RoBERTa &           42.70 (374) &           43.23 (356) &           58.01 (361) & 61.86 \\[1ex]
\multirow{3}{*}{\rotatebox[origin=c]{90}{\parbox[c]{3em}{\centering \textbf{None}}}}
& BERT    &           38.85 (346) &           37.80 (330) &           54.60 (314) & 55.33 \\
& DeBERTa & \bfseries 46.16 (409) & \bfseries 47.01 (407) & \bfseries 60.43 (382) & 64.16 \\
& RoBERTa &           44.69 (400) &           44.72 (395) &           59.51 (375) & \bfseries 65.13 \\
\midrule
\multirow{3}{*}{\rotatebox[origin=c]{90}{\parbox[c]{3em}{\centering \textbf{Oracle}}}}
& BERT    &           40.69 (367) &           39.02 (366) &           58.05 (324) & 54.84 \\
& DeBERTa & \bfseries 50.58 (457) & \bfseries 49.40 (440) & \bfseries 64.36 (411) & \bfseries 65.50\\
& RoBERTa &           48.10 (438) &           48.57 (438) &           62.71 (400) & 63.44 \\
\bottomrule
\end{tabular}
\end{table}

%% file: acl22-clickbait-spoiling-sum.tex
\section{Conclusion}
\label{conclusion}

Clickbait spoiling is a new task to help social media users who do not want to be manipulated into falling for clickbait links. Unlike clickbait detection, which often involves filtering out clickbait posts from users' timelines, clickbait spoiling subverts the curiosity triggered by clickbait, presenting users with the withheld ``punchline'' in advance.

We compile the first large resource for clickbait with associated spoilers. By interpreting clickbait spoiling as either a question answering task or a passage retrieval task, many possible approaches are available to extract from the linked document of a clickbait post the phrase or passage that spoils it. We have explored the effectiveness of a number of state-of-the-art solutions for both tasks in a large-scale experiment, including fine-tuning the respective models on our resource to determine their effectiveness for type-specific clickbait spoiling. Our experimental setup considers type-specific spoiling on the one hand, but on the other hand it also includes an end-to-end configuration for comparison. Overall, our results show that type-agnostic question answering-based spoiling is the most effective yet, but also that spoiler type-specific solutions have the potential to outperform them.

In addition to the possibilities explored, there might also be other approaches to clickbait spoiling: for example, paraphrasing technology could be used to directly transform a clickbait post into a version that contains its own spoiler. With respect to multipart spoilers, the use of summarization models could be an interesting direction to select the different parts of the linked document of a clickbait post that make up its multipart spoiler.

\section*{Acknowledgement}

We thank Tim Gollub, and our students Jana Puschmann and Bagrat Ter-Akopyan, who helped to create earlier versions of the dataset.

%% file: acl22-clickbait-spoiling-ethics.tex
\section*{Ethics Statement}

The spread of clickbait on social media by news publishers to promote click-through to their websites has been empirically found to decrease their perceived credibility in readers \cite{molyneux:2020}. There is, of course, nothing wrong with monitoring and optimizing the effectiveness of marketing a newly published news article, especially in cases where the editors make an honest effort to reach and inform their target audience. But the clickbait in our corpus mostly spreads trivial facts that could have been easily fitted into the length limits of a social media post, which is why we consider these posts to fall short of the journalistic ideal. However, it is as of yet unclear, in terms of journalism ethics, whether clickbait is an acceptable means to an end for publishers (i.e., whether it is ``necessary in driving audiences to the journalism they need by giving them the journalism they seem to want.''), or whether it serves to ``crowding out \guillemotleft{}real\guillemotright{} journalism by reducing quality in favor of the need for a click-through at whatever cost'' \cite{harte:2021}.

Facebook intervened twice with algorithmic filters to reduce the amount of clickbait that people are exposed to in their timelines---even though this probably also lowered Facebook's user engagement metrics. Our technology demonstrates another, complementary way of relatively simply circumventing the purported exploitation of the curiosity gap by giving the audience a choice on whether or not they wish their cognitive ``loopholes'' to be exploited. If a sufficiently large portion of people decide to adopt spoiling tools, that would send a clear message to publishers and social media platforms alike. Spoiling clickbait, as opposed to removing it, however, still gives publishers the benefit of the doubt, since, as the publishers claim, there are people who enjoy these kinds of trivia.

%% file: acl22-clickbait-spoiling-frame.bbl
\begin{thebibliography}{46}
\expandafter\ifx\csname natexlab\endcsname\relax\def\natexlab#1{#1}\fi

\bibitem[{Banerjee and Lavie(2005)}]{banerjee:2005}
Satanjeev Banerjee and Alon Lavie. 2005.
\newblock \href {https://aclanthology.org/W05-0909/} {{{METEOR:} An Automatic
  Metric for {MT} Evaluation with Improved Correlation with Human Judgments}}.
\newblock In \emph{Proceedings of the Workshop on Intrinsic and Extrinsic
  Evaluation Measures for Machine Translation and/or Summarization@ACL 2005,
  Ann Arbor, Michigan, USA, June 29, 2005}, pages 65--72. Association for
  Computational Linguistics.

\bibitem[{Chakraborty et~al.(2016)Chakraborty, Paranjape, Kakarla, and
  Ganguly}]{chakraborty:2016}
Abhijnan Chakraborty, Bhargavi Paranjape, Sourya Kakarla, and Niloy Ganguly.
  2016.
\newblock \href {https://doi.org/10.1109/ASONAM.2016.7752207} {{Stop Clickbait:
  Detecting and Preventing Clickbaits in Online News Media}}.
\newblock In \emph{2016 {IEEE/ACM} International Conference on Advances in
  Social Networks Analysis and Mining, {ASONAM} 2016, San Francisco, CA, USA,
  August 18-21, 2016}, pages 9--16.

\bibitem[{Clark and Gardner(2018)}]{clark:2018}
Christopher Clark and Matt Gardner. 2018.
\newblock \href {https://doi.org/10.18653/v1/P18-1078} {{Simple and Effective
  Multi-Paragraph Reading Comprehension}}.
\newblock In \emph{Proceedings of the 56th Annual Meeting of the Association
  for Computational Linguistics, {ACL} 2018, Melbourne, Australia, July 15-20,
  2018, Volume 1: Long Papers}, pages 845--855. Association for Computational
  Linguistics.

\bibitem[{Clark et~al.(2020)Clark, Luong, Le, and Manning}]{clark:2020}
Kevin Clark, Minh{-}Thang Luong, Quoc~V. Le, and Christopher~D. Manning. 2020.
\newblock \href {https://openreview.net/forum?id=r1xMH1BtvB} {{ELECTRA:}
  pre-training text encoders as discriminators rather than generators}.
\newblock In \emph{8th International Conference on Learning Representations,
  {ICLR} 2020, Addis Ababa, Ethiopia, April 26-30, 2020}. OpenReview.net.

\bibitem[{Craswell et~al.(2020)Craswell, Mitra, Yilmaz, and
  Campos}]{craswell:2020}
Nick Craswell, Bhaskar Mitra, Emine Yilmaz, and Daniel Campos. 2020.
\newblock \href {https://trec.nist.gov/pubs/trec29/papers/OVERVIEW.DL.pdf}
  {{Overview of the {TREC} 2020 Deep Learning Track}}.
\newblock In \emph{Proceedings of the Twenty-Ninth Text REtrieval Conference,
  {TREC} 2020, Virtual Event [Gaithersburg, Maryland, USA], November 16-20,
  2020}, volume 1266 of \emph{{NIST} Special Publication}. National Institute
  of Standards and Technology {(NIST)}.

\bibitem[{Craswell et~al.(2019)Craswell, Mitra, Yilmaz, Campos, and
  Voorhees}]{craswell:2019}
Nick Craswell, Bhaskar Mitra, Emine Yilmaz, Daniel Campos, and Ellen~M.
  Voorhees. 2019.
\newblock \href {https://trec.nist.gov/pubs/trec28/papers/OVERVIEW.DL.pdf}
  {{Overview of the {TREC} 2019 Deep Learning Track}}.
\newblock In \emph{Proceedings of the Twenty-Eighth Text REtrieval Conference,
  {TREC} 2019, Gaithersburg, Maryland, USA, November 13-15, 2019}, volume 1250
  of \emph{{NIST} Special Publication}. National Institute of Standards and
  Technology {(NIST)}.

\bibitem[{Dai et~al.(2020)Dai, Lai, Yang, and Le}]{dai:2020}
Zihang Dai, Guokun Lai, Yiming Yang, and Quoc Le. 2020.
\newblock \href
  {https://proceedings.neurips.cc/paper/2020/hash/2cd2915e69546904e4e5d4a2ac9e1652-Abstract.html}
  {{Funnel-Transformer: Filtering out Sequential Redundancy for Efficient
  Language Processing}}.
\newblock In \emph{Advances in Neural Information Processing Systems 33: Annual
  Conference on Neural Information Processing Systems 2020, NeurIPS 2020,
  December 6-12, 2020, virtual}.

\bibitem[{Denkowski and Lavie(2014)}]{denkowski:2014}
{Michael J.} Denkowski and Alon Lavie. 2014.
\newblock \href {https://doi.org/10.3115/v1/w14-3348} {{Meteor Universal:
  Language Specific Translation Evaluation for Any Target Language}}.
\newblock In \emph{Proceedings of the Ninth Workshop on Statistical Machine
  Translation, WMT@ACL 2014, June 26-27, 2014, Baltimore, Maryland, {USA}},
  pages 376--380. The Association for Computer Linguistics.

\bibitem[{Devlin et~al.(2019)Devlin, Chang, Lee, and Toutanova}]{devlin:2019}
Jacob Devlin, Ming{-}Wei Chang, Kenton Lee, and Kristina Toutanova. 2019.
\newblock \href {https://doi.org/10.18653/v1/n19-1423} {{{BERT:} Pre-training
  of Deep Bidirectional Transformers for Language Understanding}}.
\newblock In \emph{Proceedings of the 2019 Conference of the North American
  Chapter of the Association for Computational Linguistics: Human Language
  Technologies, {NAACL-HLT} 2019, Minneapolis, MN, USA, June 2-7, 2019, Volume
  1 (Long and Short Papers)}, pages 4171--4186. Association for Computational
  Linguistics.

\bibitem[{Dzendzik et~al.(2021)Dzendzik, Foster, and Vogel}]{dzendzik:2021}
Daria Dzendzik, Jennifer Foster, and Carl Vogel. 2021.
\newblock \href {https://doi.org/10.18653/v1/2021.emnlp-main.693} {{English
  Machine Reading Comprehension Datasets: {A} Survey}}.
\newblock In \emph{Proceedings of the 2021 Conference on Empirical Methods in
  Natural Language Processing, {EMNLP} 2021, Virtual Event / Punta Cana,
  Dominican Republic, 7-11 November, 2021}, pages 8784--8804. Association for
  Computational Linguistics.

\bibitem[{Eidnes(2015)}]{eidnes:2015}
Lars Eidnes. 2015.
\newblock {Auto-Generating Clickbait With Recurrent Neural Networks}.
\newblock
  \href{https://web.archive.org/web/20220223161935/http://larseidnes.com/2015/10/13/auto-generating-clickbait-with-recurrent-neural-networks/}{https://web.archive.org\\/web/20220223161935/http://larseidnes.com/2015/\\10/13/auto-generating-clickbait-with-recurrent-\\neural-networks/}.

\bibitem[{{El-Arini} and Tang(2014)}]{elarini:2014}
Khalid {El-Arini} and Joyce Tang. 2014.
\newblock {News Feed FYI: Click-baiting}.
\newblock
  \href{http://web.archive.org/web/20150529104738/http://newsroom.fb.com/news/2014/08/news-feed-fyi-click-baiting/}{http://web.archive.org/web/2015052\\9104738/http://newsroom.fb.com/news/2014/08/\\news-feed-fyi-click-baiting/}.

\bibitem[{Gokaslan and Cohen(2019)}]{gokaslan:2019}
Aaron Gokaslan and Vanya Cohen. 2019.
\newblock {OpenWebText Corpus}.
\newblock
  \href{http://Skylion007.github.io/OpenWebTextCorpus}{http://Skylion007.github.io/OpenWeb\\TextCorpus}.

\bibitem[{Guo et~al.(2020)Guo, Fan, Pang, Yang, Ai, Zamani, Wu, Croft, and
  Cheng}]{guo:2020}
Jiafeng Guo, Yixing Fan, Liang Pang, Liu Yang, Qingyao Ai, Hamed Zamani, Chen
  Wu, W.~Bruce Croft, and Xueqi Cheng. 2020.
\newblock \href {https://doi.org/10.1016/j.ipm.2019.102067} {{A Deep Look Into
  Neural Ranking Models for Information Retrieval}}.
\newblock \emph{Inf. Process. Manag.}, 57(6):102067.

\bibitem[{Harte(2021)}]{harte:2021}
David Harte. 2021.
\newblock \href
  {https://www.taylorfrancis.com/chapters/edit/10.4324/9780429262708-45/clickbait-banal-news-david-harte}
  {{Clickbait and Banal News}}.
\newblock In \emph{The Routledge Companion to Journalism Ethics}, pages
  346--353. Routledge.

\bibitem[{He et~al.(2021)He, Liu, Gao, and Chen}]{he:2020}
Pengcheng He, Xiaodong Liu, Jianfeng Gao, and Weizhu Chen. 2021.
\newblock \href {https://openreview.net/forum?id=XPZIaotutsD} {{DeBERTa:
  {D}ecoding-Enhanced {BERT} with Disentangled Attention}}.
\newblock In \emph{9th International Conference on Learning Representations,
  {ICLR} 2021, Virtual Event, Austria, May 3-7, 2021}. OpenReview.net.

\bibitem[{Joshi et~al.(2017)Joshi, Choi, Weld, and Zettlemoyer}]{joshi:2017}
Mandar Joshi, Eunsol Choi, Daniel~S. Weld, and Luke Zettlemoyer. 2017.
\newblock \href {https://doi.org/10.18653/v1/P17-1147} {{TriviaQA: {A} Large
  Scale Distantly Supervised Challenge Dataset for Reading Comprehension}}.
\newblock In \emph{Proceedings of the 55th Annual Meeting of the Association
  for Computational Linguistics, {ACL} 2017, Vancouver, Canada, July 30 -
  August 4, Volume 1: Long Papers}, pages 1601--1611. Association for
  Computational Linguistics.

\bibitem[{Karn et~al.(2019)Karn, Buckley, Waltinger, and
  Sch{\"{u}}tze}]{karn:2019}
{Sanjeev Kumar} Karn, Mark Buckley, Ulli Waltinger, and Hinrich Sch{\"{u}}tze.
  2019.
\newblock \href {https://doi.org/10.18653/v1/n19-1398} {{News Article Teaser
  Tweets and How to Generate Them}}.
\newblock In \emph{Proceedings of the 2019 Conference of the North American
  Chapter of the Association for Computational Linguistics: Human Language
  Technologies, {NAACL-HLT} 2019, Minneapolis, MN, USA, June 2-7, 2019, Volume
  1 (Long and Short Papers)}, pages 3967--3977. Association for Computational
  Linguistics.

\bibitem[{Koechley(2012)}]{koechley:2012}
Peter Koechley. 2012.
\newblock {Why The Title Matters More Than The Talk}.
\newblock
  \href{http://web.archive.org/web/20150611110506/http://blog.upworthy.com/post/26345634089/why-the-title-matters-more-than-the-talk}{http://web.archive.org/web/2015061\\1110506/http://blog.upworthy.com/post/263456340\\89/why-the-title-matters-more-than-the-talk}.

\bibitem[{Lan et~al.(2020)Lan, Chen, Goodman, Gimpel, Sharma, and
  Soricut}]{lan:2019}
Zhenzhong Lan, Mingda Chen, Sebastian Goodman, Kevin Gimpel, Piyush Sharma, and
  Radu Soricut. 2020.
\newblock \href {https://openreview.net/forum?id=H1eA7AEtvS} {{{ALBERT:} {A}
  Lite {BERT} for Self-supervised Learning of Language Representations}}.
\newblock In \emph{8th International Conference on Learning Representations,
  {ICLR} 2020, Addis Ababa, Ethiopia, April 26-30, 2020}. OpenReview.net.

\bibitem[{Lin et~al.(2021)Lin, Nogueira, and Yates}]{lin:2020}
Jimmy Lin, Rodrigo Nogueira, and Andrew Yates. 2021.
\newblock \href {https://doi.org/10.2200/S01123ED1V01Y202108HLT053}
  {\emph{{Pretrained Transformers for Text Ranking: {BERT} and Beyond}}}.
\newblock Synthesis Lectures on Human Language Technologies. Morgan {\&}
  Claypool Publishers.

\bibitem[{Lischka and Garz(2021)}]{lischka:2021}
{Juliane A} Lischka and Marcel Garz. 2021.
\newblock \href {https://doi.org/10.1177/14614448211027174} {{Clickbait News
  and Algorithmic Curation: A Game Theory Framework of the Relation between
  Journalism, Users, and Platforms}}.
\newblock \emph{New Media \& Society}.

\bibitem[{Liu et~al.(2019)Liu, Ott, Goyal, Du, Joshi, Chen, Levy, Lewis,
  Zettlemoyer, and Stoyanov}]{liu:2019}
Yinhan Liu, Myle Ott, Naman Goyal, Jingfei Du, Mandar Joshi, Danqi Chen, Omer
  Levy, Mike Lewis, Luke Zettlemoyer, and Veselin Stoyanov. 2019.
\newblock \href {http://arxiv.org/abs/1907.11692} {{RoBERTa: {A} Robustly
  Optimized {BERT} Pretraining Approach}}.
\newblock \emph{CoRR}, abs/1907.11692.

\bibitem[{Loewenstein(1994)}]{loewenstein:1994}
George Loewenstein. 1994.
\newblock \href
  {https://www.researchgate.net/publication/232440476_The_Psychology_of_Curiosity_A_Review_and_Reinterpretation}
  {{The Psychology of Curiosity: A Review and Reinterpretation}}.
\newblock \emph{Psychological Bulletin}, 116(1):75--98.

\bibitem[{Molyneux and Coddington(2020)}]{molyneux:2020}
Logan Molyneux and Mark Coddington. 2020.
\newblock \href {https://doi.org/10.1080/17512786.2019.1628658} {{Aggregation,
  Clickbait and Their Effect on Perceptions of Journalistic Credibility and
  Quality}}.
\newblock \emph{Journalism Practice}, 14(4):429--446.

\bibitem[{Munger(2020)}]{munger:2020}
Kevin Munger. 2020.
\newblock \href {https://doi.org/10.1080/10584609.2019.1687626} {{All the News
  That’s Fit to Click: The Economics of Clickbait Media}}.
\newblock \emph{Political Communication}, 37(3):376--397.

\bibitem[{Nguyen et~al.(2016)Nguyen, Rosenberg, Song, Gao, Tiwary, Majumder,
  and Deng}]{nguyen:2016}
Tri Nguyen, Mir Rosenberg, Xia Song, Jianfeng Gao, Saurabh Tiwary, Rangan
  Majumder, and Li~Deng. 2016.
\newblock \href {http://ceur-ws.org/Vol-1773/CoCoNIPS\_2016\_paper9.pdf} {{{MS}
  {MARCO:} {A} Human Generated MAchine Reading COmprehension Dataset}}.
\newblock In \emph{Proceedings of the Workshop on Cognitive Computation:
  Integrating neural and symbolic approaches 2016 co-located with the 30th
  Annual Conference on Neural Information Processing Systems {(NIPS} 2016),
  Barcelona, Spain, December 9, 2016}, volume 1773 of \emph{{CEUR} Workshop
  Proceedings}. CEUR-WS.org.

\bibitem[{Nogueira and Cho(2019)}]{nogueira:2019a}
Rodrigo Nogueira and Kyunghyun Cho. 2019.
\newblock \href {http://arxiv.org/abs/1901.04085} {{Passage Re-ranking with
  {BERT}}}.
\newblock \emph{CoRR}, abs/1901.04085.

\bibitem[{Nogueira et~al.(2020)Nogueira, Jiang, Pradeep, and
  Lin}]{nogueira:2020}
Rodrigo Nogueira, Zhiying Jiang, Ronak Pradeep, and Jimmy Lin. 2020.
\newblock \href {https://doi.org/10.18653/v1/2020.findings-emnlp.63} {{Document
  Ranking with a Pretrained Sequence-to-Sequence Model}}.
\newblock In \emph{Findings of the Association for Computational Linguistics:
  {EMNLP} 2020, Online Event, 16-20 November 2020}, volume {EMNLP} 2020 of
  \emph{Findings of {ACL}}, pages 708--718. Association for Computational
  Linguistics.

\bibitem[{Nogueira et~al.(2019)Nogueira, Yang, Cho, and Lin}]{nogueira:2019b}
Rodrigo Nogueira, Wei Yang, Kyunghyun Cho, and Jimmy Lin. 2019.
\newblock \href {http://arxiv.org/abs/1910.14424} {{Multi-Stage Document
  Ranking with {BERT}}}.
\newblock \emph{CoRR}, abs/1910.14424.

\bibitem[{Papineni et~al.(2002)Papineni, Roukos, Ward, and Zhu}]{papineni:2002}
Kishore Papineni, Salim Roukos, Todd Ward, and Wei{-}Jing Zhu. 2002.
\newblock \href {https://doi.org/10.3115/1073083.1073135} {{{BLEU}: {A} Method
  for Automatic Evaluation of Machine Translation}}.
\newblock In \emph{Proceedings of the 40th Annual Meeting of the Association
  for Computational Linguistics, July 6-12, 2002, Philadelphia, PA, {USA}},
  pages 311--318. {ACL}.

\bibitem[{Peysakhovich and Hendrix(2016)}]{peysakhovich:2016}
Alex Peysakhovich and Kristin Hendrix. 2016.
\newblock {Further Reducing Clickbait in Feed}.
\newblock
  \href{https://web.archive.org/web/20210207042429/https://about.fb.com/news/2016/08/news-feed-fyi-further-reducing-clickbait-in-feed/}{https://web.archive.org/\\web/20210207042429/https://about.fb.com/news/20\\16/08/news-feed-fyi-further-reducing-clickbait-in-\\feed/}.

\bibitem[{Ponte and Croft(1998)}]{ponte:1998}
{Jay M.} Ponte and {W. Bruce} Croft. 1998.
\newblock \href {https://doi.org/10.1145/290941.291008} {{A Language Modeling
  Approach to Information Retrieval}}.
\newblock In \emph{{SIGIR} '98: Proceedings of the 21st Annual International
  {ACM} {SIGIR} Conference on Research and Development in Information
  Retrieval, August 24-28 1998, Melbourne, Australia}, pages 275--281. {ACM}.

\bibitem[{Potthast et~al.(2018)Potthast, Gollub, Hagen, and
  Stein}]{potthast:2018w}
Martin Potthast, Tim Gollub, Matthias Hagen, and Benno Stein. 2018.
\newblock \href {https://arxiv.org/abs/1812.10847} {{The Clickbait Challenge
  2017: Towards a Regression Model for Clickbait Strength}}.
\newblock \emph{CoRR}, abs/1812.10847.

\bibitem[{Potthast et~al.(2016)Potthast, K{\"o}psel, Stein, and
  Hagen}]{potthast:2016b}
Martin Potthast, Sebastian K{\"o}psel, Benno Stein, and Matthias Hagen. 2016.
\newblock \href {https://doi.org/10.1007/978-3-319-30671-1\_72} {{Clickbait
  Detection}}.
\newblock In \emph{Advances in Information Retrieval. 38th European Conference
  on IR Research (ECIR 2016)}, volume 9626 of \emph{Lecture Notes in Computer
  Science}, pages 810--817, Berlin Heidelberg New York. Springer.

\bibitem[{Rajpurkar et~al.(2018)Rajpurkar, Jia, and Liang}]{rajpurkar:2018}
Pranav Rajpurkar, Robin Jia, and Percy Liang. 2018.
\newblock \href {https://doi.org/10.18653/v1/P18-2124} {{Know What You Don't
  Know: Unanswerable Questions for SQuAD}}.
\newblock In \emph{Proceedings of the 56th Annual Meeting of the Association
  for Computational Linguistics, {ACL} 2018, 2: Short Papers}, pages 784--789.
  Association for Computational Linguistics.

\bibitem[{Rajpurkar et~al.(2016)Rajpurkar, Zhang, Lopyrev, and
  Liang}]{rajpurkar:2016}
Pranav Rajpurkar, Jian Zhang, Konstantin Lopyrev, and Percy Liang. 2016.
\newblock \href {https://doi.org/10.18653/v1/d16-1264} {{SQuAD: 100,000+
  Questions for Machine Comprehension of Text}}.
\newblock In \emph{Proceedings of the 2016 Conference on Empirical Methods in
  Natural Language Processing, {EMNLP} 2016, Austin, Texas, USA, November 1-4,
  2016}, pages 2383--2392. The Association for Computational Linguistics.

\bibitem[{Robertson and Zaragoza(2009)}]{robertson:2009}
Stephen~E. Robertson and Hugo Zaragoza. 2009.
\newblock \href {https://doi.org/10.1561/1500000019} {{The Probabilistic
  Relevance Framework: {BM25} and Beyond}}.
\newblock \emph{Found. Trends Inf. Retr.}, 3(4):333--389.

\bibitem[{Rony et~al.(2017)Rony, Hassan, and Yousuf}]{rony:2017}
{Md M. U.} Rony, Naeemul Hassan, and Mohammad Yousuf. 2017.
\newblock \href {https://doi.org/10.1145/3110025.3110054} {{Diving Deep into
  Clickbaits: Who Use Them to What Extents in Which Topics with What Effects?}}
\newblock In \emph{Proceedings of the 2017 {IEEE/ACM} International Conference
  on Advances in Social Networks Analysis and Mining 2017, Sydney, Australia,
  July 31 - August 03, 2017}, pages 232--239. {ACM}.

\bibitem[{Rubin et~al.(2015)Rubin, Conroy, and Chen}]{rubin:2015a}
Victoria Rubin, Niall Conroy, and Yimin Chen. 2015.
\newblock \href
  {http://socsci-dev.ss.uci.edu/~lpearl/courses/readings/RubinEtAl2015_DeceptionDetectionNews.pdf}
  {{Towards News Verification: Deception Detection Methods for News
  Discourse}}.
\newblock In \emph{Proceedings of the Hawaii International Conference on System
  Sciences (HICSS48) Symposium on Rapid Screening Technologies, Deception
  Detection and Credibility Assessment Symposium}, Kauai, Hawaii, USA.

\bibitem[{Shu et~al.(2018)Shu, Wang, Le, Lee, and Liu}]{shu:2018}
Kai Shu, Suhang Wang, Thai Le, Dongwon Lee, and Huan Liu. 2018.
\newblock \href {https://doi.org/10.1109/ICDM.2018.00062} {{Deep Headline
  Generation for Clickbait Detection}}.
\newblock In \emph{{IEEE} International Conference on Data Mining, {ICDM} 2018,
  Singapore, November 17-20, 2018}, pages 467--476. {IEEE} Computer Society.

\bibitem[{Song et~al.(2020)Song, Tan, Qin, Lu, and Liu}]{song:2020}
Kaitao Song, Xu~Tan, Tao Qin, Jianfeng Lu, and Tie{-}Yan Liu. 2020.
\newblock \href
  {https://proceedings.neurips.cc/paper/2020/hash/c3a690be93aa602ee2dc0ccab5b7b67e-Abstract.html}
  {{MPNet: Masked and Permuted Pre-training for Language Understanding}}.
\newblock In \emph{Advances in Neural Information Processing Systems 33: Annual
  Conference on Neural Information Processing Systems 2020, NeurIPS 2020,
  December 6-12, 2020, virtual}.

\bibitem[{Xu et~al.(2019)Xu, Wu, Madotto, and Fung}]{xu:2019}
Peng Xu, Chien{-}Sheng Wu, Andrea Madotto, and Pascale Fung. 2019.
\newblock \href {https://doi.org/10.18653/v1/D19-1303} {{Clickbait? Sensational
  Headline Generation with Auto-tuned Reinforcement Learning}}.
\newblock In \emph{Proceedings of the 2019 Conference on Empirical Methods in
  Natural Language Processing and the 9th International Joint Conference on
  Natural Language Processing, {EMNLP-IJCNLP} 2019, Hong Kong, China, November
  3-7, 2019}, pages 3063--3073. Association for Computational Linguistics.

\bibitem[{Yang et~al.(2017)Yang, Fang, and Lin}]{yang:2017}
Peilin Yang, Hui Fang, and Jimmy Lin. 2017.
\newblock \href {https://doi.org/10.1145/3077136.3080721} {{Anserini: Enabling
  the Use of Lucene for Information Retrieval Research}}.
\newblock In \emph{Proceedings of the 40th International {ACM} {SIGIR}
  Conference on Research and Development in Information Retrieval, Shinjuku,
  Tokyo, Japan, August 7-11, 2017}, pages 1253--1256. {ACM}.

\bibitem[{Zaheer et~al.(2020)Zaheer, Guruganesh, Dubey, Ainslie, Alberti,
  Onta{\~{n}}{\'{o}}n, Pham, Ravula, Wang, Yang, and Ahmed}]{zaheer:2020}
Manzil Zaheer, Guru Guruganesh, Kumar~Avinava Dubey, Joshua Ainslie, Chris
  Alberti, Santiago Onta{\~{n}}{\'{o}}n, Philip Pham, Anirudh Ravula, Qifan
  Wang, Li~Yang, and Amr Ahmed. 2020.
\newblock \href
  {https://proceedings.neurips.cc/paper/2020/hash/c8512d142a2d849725f31a9a7a361ab9-Abstract.html}
  {{Big Bird: Transformers for Longer Sequences}}.
\newblock In \emph{Advances in Neural Information Processing Systems 33: Annual
  Conference on Neural Information Processing Systems 2020, NeurIPS 2020,
  December 6-12, 2020, virtual}.

\bibitem[{Zhang et~al.(2020)Zhang, Kishore, Wu, Weinberger, and
  Artzi}]{zhang:2020}
Tianyi Zhang, Varsha Kishore, Felix Wu, Kilian~Q. Weinberger, and Yoav Artzi.
  2020.
\newblock \href {https://openreview.net/forum?id=SkeHuCVFDr} {{BERTScore:
  Evaluating Text Generation with {BERT}}}.
\newblock In \emph{8th International Conference on Learning Representations,
  {ICLR} 2020, Addis Ababa, Ethiopia, April 26-30, 2020}. OpenReview.net.

\end{thebibliography}
